\documentclass[letterpaper, 10 pt, conference]{ieeeconf}  

\IEEEoverridecommandlockouts                              

\usepackage{graphicx}
\usepackage{amsmath}
\usepackage{amssymb}
\usepackage{booktabs}
\usepackage{subcaption}
\usepackage{caption}

\usepackage[dvipsnames]{xcolor}
\usepackage[colorlinks=true, linkcolor=blue, urlcolor=blue, citecolor=blue]{hyperref}

\usepackage{lipsum}
\usepackage{afterpage}

\makeatletter
\newcommand{\setcaptype}[1]{\def\@captype{#1}}
\makeatother

\newsavebox{\tempbox}

\title
{\LARGE \bf
Aug3D: Augmenting large scale outdoor datasets for Generalizable Novel View Synthesis\\
{\small \href{https://aug3Dim.github.io}{\texttt{aug3Dim.github.io}}}%
}

\author{Aditya Rauniyar*$^{1}$, Omar Alama*$^{1}$, Silong Yong$^{1}$, Katia Sycara$^{1}$, and Sebastian Scherer$^{1}$
\thanks{*The authors share equal contribution.}
\thanks{$^{1}$A. Rauniyar, O. Alama, S. Yong, K.Sycara and S. Scherer are with the Robotics Institute, School of Computer Science at Carnegie Mellon University, Pittsburgh, PA, USA.
       {\tt\small \{arauniya, oalama, silongy, sycara, basti\}@andrew.cmu.edu}}
\thanks{This work is supported by the Defense Science and Technology Agency Singapore contract DST000EC124000205.}%
}

\begin{document}


\maketitle

\savebox{\tempbox}{\begin{minipage}{\textwidth}
\setcaptype{figure}%
\centering
    \includegraphics[width=\textwidth,height=\textheight,keepaspectratio]{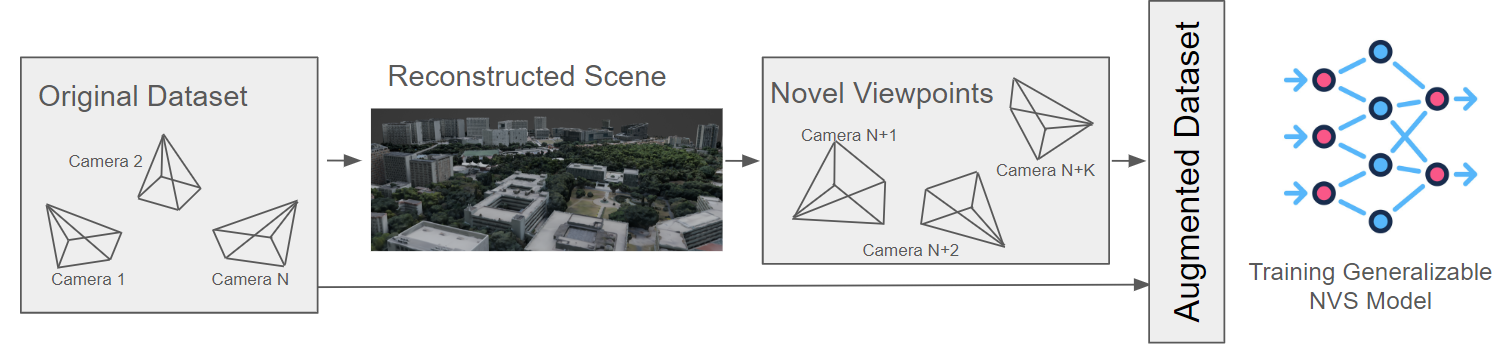}
    \caption{Aug3D addresses challenges with low-overlap clusters in large-scale outdoor datasets for generalizable novel view synthesis by reconstructing scenes, sampling camera poses to mitigate overlap issues, and combining these samples with the real dataset, resulting in improved performance.}

    \label{fig:aug3d}
\end{minipage}}

\begin{figure}[t]
\rlap{\usebox\tempbox}
\end{figure}
\afterpage{\begin{figure}[t]
\rule{0pt}{\dimexpr \ht\tempbox+\dp\tempbox}
\end{figure}}

\thispagestyle{empty}
\pagestyle{empty}

\begin{abstract}

Recent photorealistic Novel View Synthesis (NVS) advances have increasingly gained attention. 
However, these approaches remain constrained to small indoor scenes. 
While optimization-based NVS models have made attempts to address this, generalizable feed-forward methods—offering significant advantages—remain underexplored. 
In this work, we train PixelNeRF, a feed-forward NVS model, on the large-scale UrbanScene3D dataset. 
We propose four training strategies to cluster and train on this dataset, highlighting that performance is hindered by limited view overlap. 
To address this, we introduce Aug3D, an augmentation technique that leverages reconstructed scenes using traditional Structure-from-Motion (SfM). 
Aug3D generates well-conditioned novel views through grid and semantic sampling to enhance feed-forward NVS model learning.
Our experiments reveal that reducing the number of views per cluster from 20 to 10 improves PSNR by 10\%, but the performance remains suboptimal. Aug3D further addresses this by combining the newly generated novel views with the original dataset, demonstrating its effectiveness in improving the model's ability to predict novel views.

\end{abstract}


\section{Introduction}
\label{sec:intro}


Photorealistic Novel View Synthesis (NVS) plays a vital role in applications requiring immersive experiences, such as AR/VR. As these methods gain popularity, there is a growing need to extend their capabilities to outdoor environments. In this study, we introduce Aug3D, a reconstruction-based augmentation technique designed to adapt existing outdoor datasets for NVS applications.   

Generalizable models, exemplified by works like PixelNeRF \cite{yu2020pixelnerf} and Splatter-Image \cite{szymanowicz_splatter_2023}, render photorealistic novel views applicable to a wider range of inputs. These models are typically trained on smaller, object-centric scenes or indoor environments. In this work, we extend the application of NVS to large outdoor environments, aiming to broaden the scope of these methods for novel view synthesis.

Alternatively, we take inspiration from scene-specific NeRF approaches in the research community, such as MegaNeRF \cite{turki_mega-nerf_2022} and VastGaussian \cite{lin2024vastgaussian}, which fine-tune the NeRF model for NVS on specific scenes. These provide insights into selecting large outdoor scenes for training generalizable models to synthesize novel views. 

\textbf{Challenges: } Utilizing large outdoor scenes for generalizing NVS models presents several hurdles. The first arises from how these scenes are typically captured using drones, often employing constant-altitude grid scans over regions of interest \cite{rematas_urban_2022, turki_mega-nerf_2022}. This results in captures that vary predominantly in a translated direction, introducing novel features to the scene between consecutive shots and posing difficulties for NVS methods to operate effectively. Additionally, most existing NVS work focuses on object-centric scene captures,  for objects or indoor/outdoor environments. Such captures are vital, as the models rely on correlated features across input images to render novel views. Furthermore, generalizable NVS models typically train on datasets with minimal variation across input images (e.g., DTU dataset \cite{jensen_large_2014} ), where input images are placed in an object-centric way and exhibit controlled changes in elevation and azimuth. As a result, novel views are interpolated rather than extrapolated. Therefore, large outdoor scene environments used for scene-specific NVS models must (1) align with existing generalizable NVS model training setups, introducing fewer new elements across input images, and (2) feature input images that are closely spaced with controlled variations in view poses (e.g., DTU \cite{jensen_large_2014}, Shapenet \cite{chang_shapenet_nodate} dataset).

\textbf{Aug3D: }To address the challenges, we introduce Aug3D \ref{fig:aug3d}, an augmentation camera sampling strategy to adapt large outdoor scene datasets such as UrbanScene3D \cite{lin_capturing_2022} and Mill-19 \cite{turki_mega-nerf_2022} for training generalizable novel view synthesis models. To mitigate sensitivity to input image poses, we cluster them into N views, maximizing shared points through Structure from Motion (SfM). However, sparse data collection via drone flight requires further measures to enhance feature correlation among input images. To accommodate poses beyond original locations and ensure scale invariance, we sample camera poses by geometric reconstruction of large scenes. While reconstruction quality impacts these views, advances in photorealistic scene-specific NVS models such as Mega-NeRF \cite{turki_mega-nerf_2022}, Block-NeRF \cite{tancik_block-nerf_2022}, and VastGaussians \cite{lin2024vastgaussian} suggest sufficient development within the research community for our proposed method.

\textbf{Contributions: } Our work addresses the question: ``How can we effectively train existing Generalizable NVS models for large-scale outdoor datasets?" 
Here are our key contributions:
\begin{itemize}
\item We cluster outdoor datasets using high point matching, aligning them with the DTU format for compatibility with any NVS model designed for the DTU dataset. We validate this approach with PixelNeRF \cite{yu2020pixelnerf}.
\item Our multi-scaling camera sampling method generates additional viewpoints not present in the original dataset. These new viewpoints, derived from mesh-based scene capture, produce synthetic renders whose quality relies on reconstruction accuracy.
\item We optimize the augmentation process with semantically aware sampling, enhancing the diversity of novel viewpoints added to the dataset. This pipeline combines geometric and feature-wise segmentation techniques.
\end{itemize}

\section{Related Work}

\textbf{Novel View Synthesis.} Novel view synthesis (NVS), tackles the challenge of synthesizing novel RGB views with a set of RGB input views (without necessarily constructing explicit 3D geometry). NVS has seen a rapid growth of interest with the advent recent breakthroughs in learning/neural based methods. These neural methods can be broadly classified into surface or volumetric based approaches\cite{Tewari2021AdvancesIN}. Neural surface approaches reason  about the surfaces in the scene either representing them implicitly with zero level set functions  \cite{zhang2021ners,Kellnhofer2021NeuralLR}, with continuous parametric methods \cite{Tulsiani2020ImplicitMR, Bhattad2021ViewGF}, or explicitly using meshes \cite{Riegler2020FreeVS}, or points/surfels \cite{Wiles2019SynSinEV,aliev2020neural}. Neural volumetric approaches reason about the volumes occupied by elements in the scene that represent them implicitly \cite{Niemeyer2019DifferentiableVR, Sitzmann2018DeepVoxelsLP}, as a Neural Radiance Field (NeRF) \cite{mildenhall_nerf_nodate}, as a set of volumetric primitives \cite{kerbl3Dgaussians}, or explicitly as voxel grids \cite{Yu2021PlenOctreesFR,fridovich2022plenoxels} or multiplane images \cite{Wizadwongsa2021NeXRV}.

In this work, we focus our evaluation on neural volumetric methods specifically, the NeRF \cite{mildenhall_nerf_nodate} lines of work due to its significant success in high fidelity novel view synthesis and the existence of efforts to extend such approaches to large-scale urban settings.

\textbf{Generalizable NVS.}
Generalizable, image-based, or feed-forward NVS refers to models that can predict novel views at test time without having to re-optimize any learnable parameters. This is done by conditioning the architecture on sets of input views and describing different scenes while training. In contrast to the optimization-based single-scene networks, feed-forward models can learn semantic priors that make them superior in sparse input NVS. 

Works like PixelNeRF \cite{yu2020pixelnerf} conditions NeRF on pixel aligned features recovered by projecting a query point onto feature maps of the input views. IBRNet \cite{wang2021ibrnet} uses a similar approach but uses transformers. MVSNeRF\cite{chen_mvsnerf_2021} uses 3D convolutions on top of a plane sweep of input images to get per voxel image features and uses that to condition NeRF per query point. MuRF\cite{xu2023murf} constructs a frustum volume aligned with the target view allowing them to utilize 3D convolutions to predict the volume. Similar recent works \cite{szymanowicz_splatter_2023, charatan_pixelsplat_2023, chen2024mvsplat} have worked on generalizing 3D Gaussian splatting through input image conditioning. 

All mentioned works focus on small to medium scale scenes with very limited target view ranges mainly due to the absence of city scale datasets amenable to feed-forward NVS. Our objective is to offer a training and data augmentation strategy to allow such works to learn large-scale urban scene priors efficiently.

\textbf{Large Scale Scene Reconstruction: } Large city-scale reconstruction has been a long-standing field of research. Many works attempt to reconstruct large scenes using traditional methods such as Lidar point clouds \cite{lan_robust_2019}, meshes \cite{valentin_mesh_2013}, or signed distance functions \cite{oleynikova_signed_2016}. However, there is an increased interest in using neural volumetric representations for their high-fidelity reconstructions. \cite{turki_mega-nerf_2022, tancik_block-nerf_2022, zhang2024birdnerf} recognize NeRF's capacity limitations and propose forms of spatial decomposition and train many NeRF's to represent different parts of the large scene. Mega-NeRF\cite{turki_mega-nerf_2022} and BirdNeRF\cite{zhang2024birdnerf} focus on bird view reconstruction, while Block-NeRF\cite{tancik_block-nerf_2022} focuses on street view. BungeeNeRF \cite{xiangli2022bungeenerf} takes a different approach focusing on satellite view reconstruction, recognizes the need for multi-scale reconstruction, and progressively trains from big to small scales while increasing network capacity. Urban Radiance Fields \cite{rematas_urban_2022} presents a multi-modal approach of combining lidar information with RGB signals to address exposure differences in outdoor scenes. VastGaussian \cite{lin2024vastgaussian} introduces spatial decomposition approaches to 3D Gaussian splatting for large-scale bird view scene reconstruction.

However, the aforementioned works develop optimization-based models that need extensive training and are unsuitable for online reconstruction during navigation or data acquisition. We explore the capabilities of feed-forward approaches to reconstruct large-scale urban scenes, allowing on the fly reconstruction times.

\textbf{Augmentation for scene understanding: } Data augmentation is a proven technique for improving ML model generalizability. Numerous augmentation methods have been developed in the 2D vision space. We take inspiration from CutOut \cite{devries_improved_2017} and CutMix \cite{yun_cutmix_2019} that cut 2D images out and mix cuts respectively. These methods however cannot be directly applied on input images for 3D NVS as they compromise cross-view consistency. Recently, 3D augmentation techniques have been developed. Notably, Mix3D \cite{nekrasov_mix3d_2021} mixes elements/meshes from different synthetic indoor scenes to compose new scenes that are not necessarily semantically reasonable to improve generalizability following the effective techniques of domain randomization\cite{tobin2017domain, tremblay2018training}. Their work however is done in a limited indoor setting for 3D semantic segmentation. There exists very few works \cite{bortolon2023vm, chen2022aug} that tackle augmentation for feed-forward NVS, they only augment in 2D image space, severely limiting the variations introduced.

\section{Approach}

\subsection{Data curation for Generalizable NVS}
\label{sec:preprocess}

\begin{figure*}[!htb]
    \centering
    \begin{subfigure}[b]{0.22\textwidth}
        \centering
        \includegraphics[width=\textwidth]{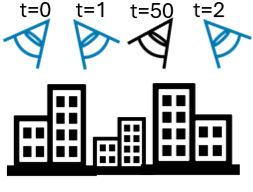}
        \caption{Capture sequence grouping}
        \label{fig:grouping_seq}
    \end{subfigure}
    \hfill
    \begin{subfigure}[b]{0.215\textwidth}
        \centering
        \includegraphics[width=\textwidth]{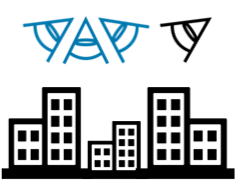}
        \caption{Grid-based grouping}
        \label{fig:grouping_grid}
    \end{subfigure}
    \hfill
    \begin{subfigure}[b]{0.22\textwidth}
        \centering
        \includegraphics[width=\textwidth]{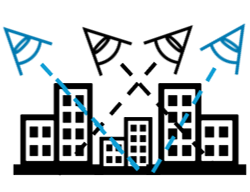}
        \caption{Ray intersection grouping}
        \label{fig:grouping_ray}
    \end{subfigure}
    \hfill
    \begin{subfigure}[b]{0.22\textwidth}
        \centering
        \includegraphics[width=\textwidth]{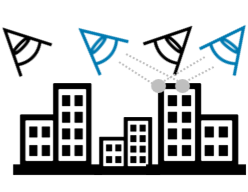}
        \caption{SfM shared points}
        \label{fig:grouping_sfm}
    \end{subfigure}
    \caption{Scene clustering methods for training GNVS models. Colored cameras represent cameras within the same cluster. (a), (b) and (c) show edge cases where these methods would cluster wrong images into a scene.}
    \label{fig:grouping_approaches_illustration}
\end{figure*}

\begin{figure*}[!htb]
    \centering
    \begin{subfigure}[b]{0.45\textwidth}
        \centering
        \includegraphics[width=\textwidth]{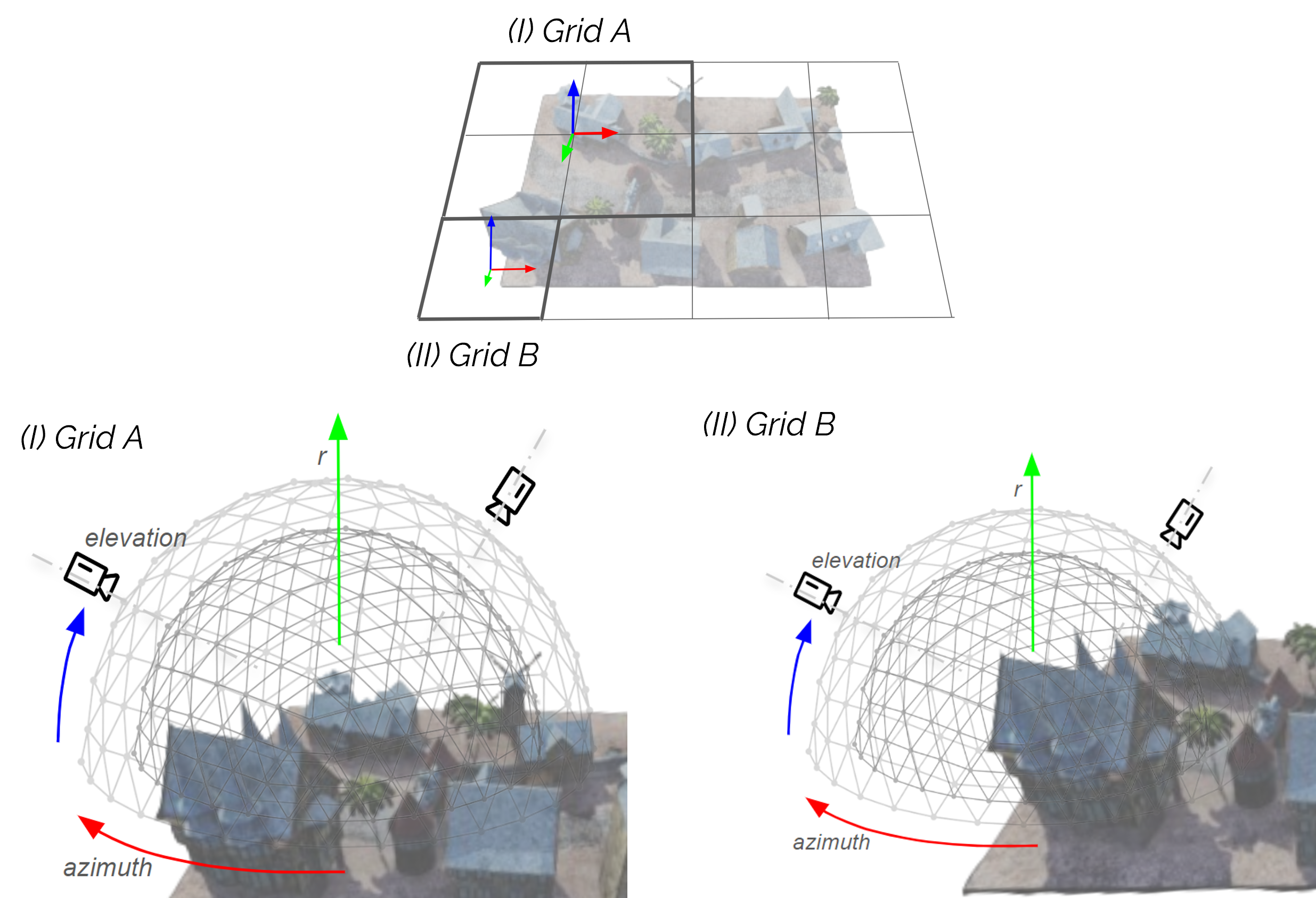}
        \caption{Multiscale Grid Sampling}
        \label{fig:grid_sampling}
    \end{subfigure}
    \hfill
    \begin{subfigure}[b]{0.52\textwidth}
        \centering
        \includegraphics[width=\textwidth]{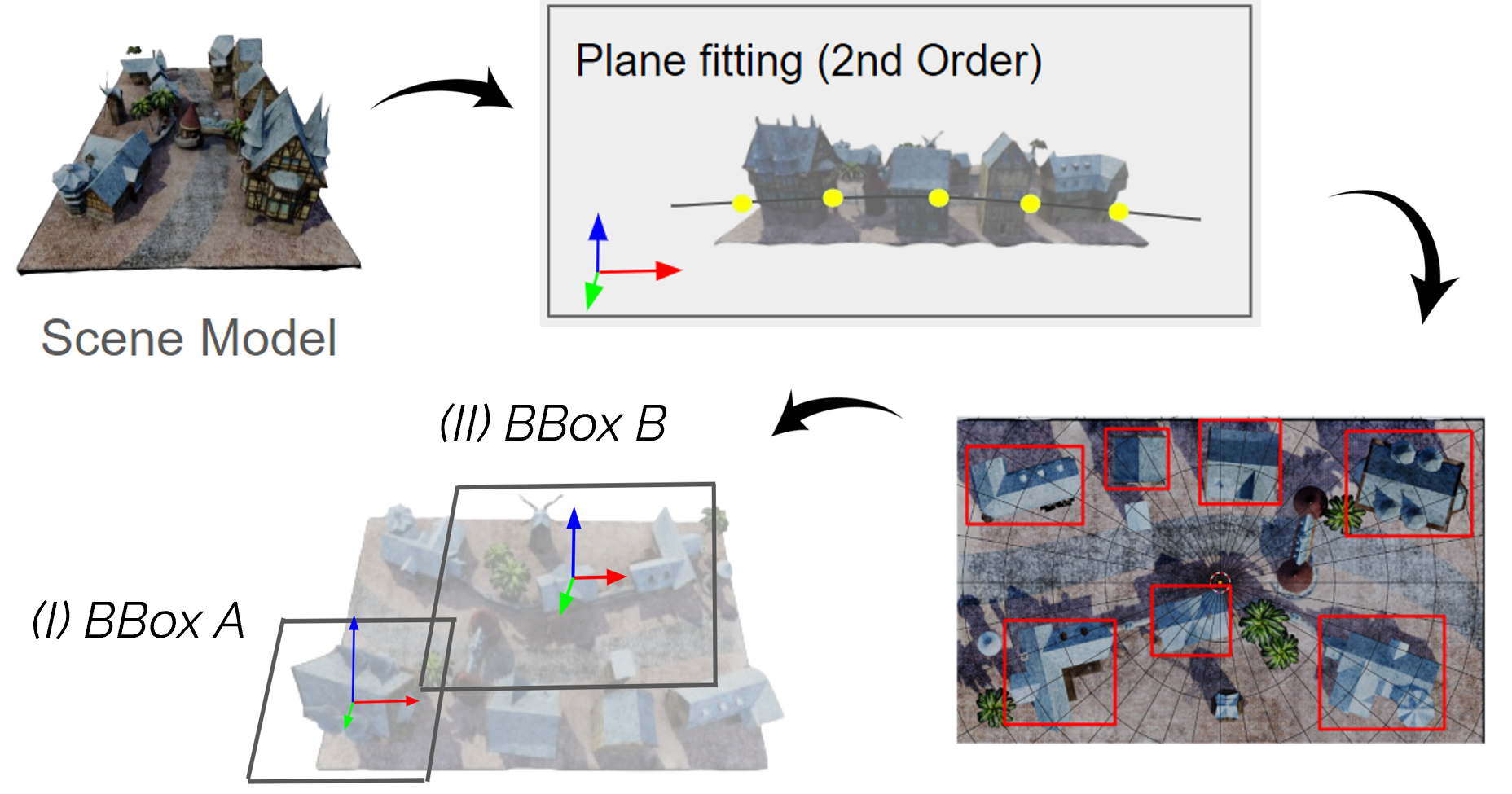}
        \caption{Semantic Sampling}
        \label{fig:semantic_sampling}
    \end{subfigure}
    \caption{Two types of augmentation to reduce low overlap among outdoor scene datasets: (a) Multiscale Grid Sampling and (b) Semantic Sampling. The left figure shows dynamic camera placements for varying grid scales, and the right figure illustrates focused sampling around urban regions.}
    \label{fig:augmentation_methods}
\end{figure*}

Large-scale urban scene data are not readily amenable for generalizable NVS as the data covers a huge baseline. For example, urbanscene3d \cite{lin_capturing_2022} real datasets can cover more than \(1 km^2\) areas spanning multiple high rise and low rise buildings. Hence, an image in the scan may not necessarily contribute meaningfully to the reconstruction of another view; clustering images meaningfully is critical. We test different algorithms as shown in Fig. \ref{fig:grouping_approaches_illustration} to achieve that targeting the following criteria:
First, it is pivotal to cluster images in the scan that are related to each other (i.e. looking more or less at the same structures in the scene). Second, the selection of the group size is crucial as too small of a group size will give very little information to the model whilst a very big group size would give confusing and unrelated information to the model. Third, the group size should be constant to allow efficient batching when training. We show a qualitative output of clustering images in Fig. \ref{fig:grouping_approaches_qualitative}. 

\subsubsection{Capture Sequence grouping} Using the capture sequence—defined as the order in which images are captured along the camera's trajectory over time—to cluster images is a straightforward but naive approach. Abrupt changes in the camera's trajectory can result in images within the same cluster capturing entirely different parts of the scene, as illustrated in Fig.\ref{fig:grouping_approaches_illustration}.a. Additionally, this method overlooks valuable images from later in the sequence that capture the same scene region but are excluded due to their temporal position.

\subsubsection{Grid-Based Grouping}
In this approach, a grid is overlaid on the ground plane, and cameras are clustered based on proximity to the centers of grid cells, with each cluster containing the K nearest neighbors to a grid cell center. While straightforward, this method has limitations: cameras that are close in Euclidean space may have vastly different viewing frustums, leading to poor clustering results, as shown in Fig.\ref{fig:grouping_approaches_illustration}.b. To address this, we added an angular constraint to ensure that cameras within a cluster are not only spatially close but also oriented in roughly the same direction. Despite this refinement, the approach still struggles to group images that capture the same scene area from different angles.

\subsubsection{Ray intersection with ground plane} 
To capture both the Euclidean distance and the viewing distance, we projected the center pixel of each image into the world frame so that it intersects with the ground plane. We then use the distances between the intersection points as our clustering metric. A drawback of this approach is that you need to estimate the distance from the ground plane to each camera. To achieve this, we use Metashape to run SfM on small hand-picked images and calculate the height of the cameras relative to the ground plane. We then use this height to calculate all other camera distances to the ground plane, assuming the ground is flat. This approach improved clustering performance but still failed in many cases near high-rise buildings, as cameras could be looking at different areas even though their rays intersect close to each other at the ground plane level.

\subsubsection{SfM shared points} To ensure that images within a cluster view the same structures, we perform a full Structure-from-Motion (SfM) process for each scene and use the number of shared points among different camera views as the metric for clustering. This approach consistently produced the best results, as illustrated in Fig. \ref{fig:grouping_approaches_qualitative}, while effectively avoiding edge cases seen in previous methods. The core idea is that cameras observing the same scene exhibit high correspondence, which we capture by computing a similarity matrix for all images in the scene using SfM. Based on this matrix, we uniformly select cluster centers across the scene and determine the top K views for each cluster according to their similarity scores. Like all clustering methods, this approach requires careful tuning of cluster size to achieve optimal performance.

\subsection{Augmentation}
\label{sec:augmentation}

Recognizing the challenges of training feed-forward NVS models directly on the real data with unconstrained capture trajectories, we further propose to augment such scenes with constrained sampling methods. First we reconstruct the scene using traditional structure from motion and multi-view stereo approaches, then sampling novel views in an object-centric manner to augment the training of the feed-forward model. We discuss various approaches to sampling in what follows below.

\textbf{Background on scene sentric dome sampling: }
A common approach for sampling images from a reconstructed scene uses an Archimedean spiral or dome above the mesh, as shown in Fig.\ref{fig:augmentation_methods}, commonly applied in models like PixelNeRF \cite{yu2020pixelnerf}. While effective for standard setups, it struggles with large scenes, often resulting in flat, disproportionate reconstructions and reliance on simple homography transformations. To address this, we propose two improved camera sampling strategies for larger scenes. 

\subsubsection{Multiscale Grid Sampling}
\label{sec:grid-based}
 
A straightforward approach involves dividing the scene into cells at varying grid scales, as shown in Fig.\ref{fig:grid_sampling}. Using multiple scales helps prevent the model from overfitting to a single scale. Virtual domes are then placed over each cell, and cameras are uniformly sampled within a limited azimuth and elevation range. To avoid manually fine-tuning grid scales for each scene, we dynamically adjust them based on the scene's height-to-width/length ratios. This ensures finer grids for large scenes and coarser grids for smaller ones, as illustrated in Fig.\ref{fig:grid_sampling}.

\subsubsection{Semantic Building Sampling}
\label{sec:semantic}
This approach focuses on underrepresented areas, such as urban regions, which are often overshadowed by forest-dominated samples. Unlike the multiscale grid method that uniformly samples the scene, this method uses semantic camera sampling to identify urban areas as regions of interest and concentrates camera samples around them. As illustrated in Fig.\ref{fig:semantic_sampling}, this strategy reduces forest overrepresentation and improves dataset diversity by prioritizing urban scenes.

\textbf{Plane fitting: }We simplify perform building detection using a geometric approach: fitting a plane to the Kth percentile of points (sorted by Z height) in the scene point cloud via least-squares. This plane slices the point cloud, rendering a top-down orthographic view, which is converted into binary masks and then bounding boxes. These bounding boxes initialize dome placements for targeted camera sampling.

To enable multiscale novel view synthesis (NVS), we extend this by combining bounding boxes. For each detected box, we merge it with 1 to M nearest boxes, creating clusters that represent individual buildings and multi-building regions, ensuring comprehensive and scalable scene coverage as showing in Fig.\ref{fig:semantic_sampling}.

\begin{figure*}
    \centering
    \includegraphics[scale = 0.6]{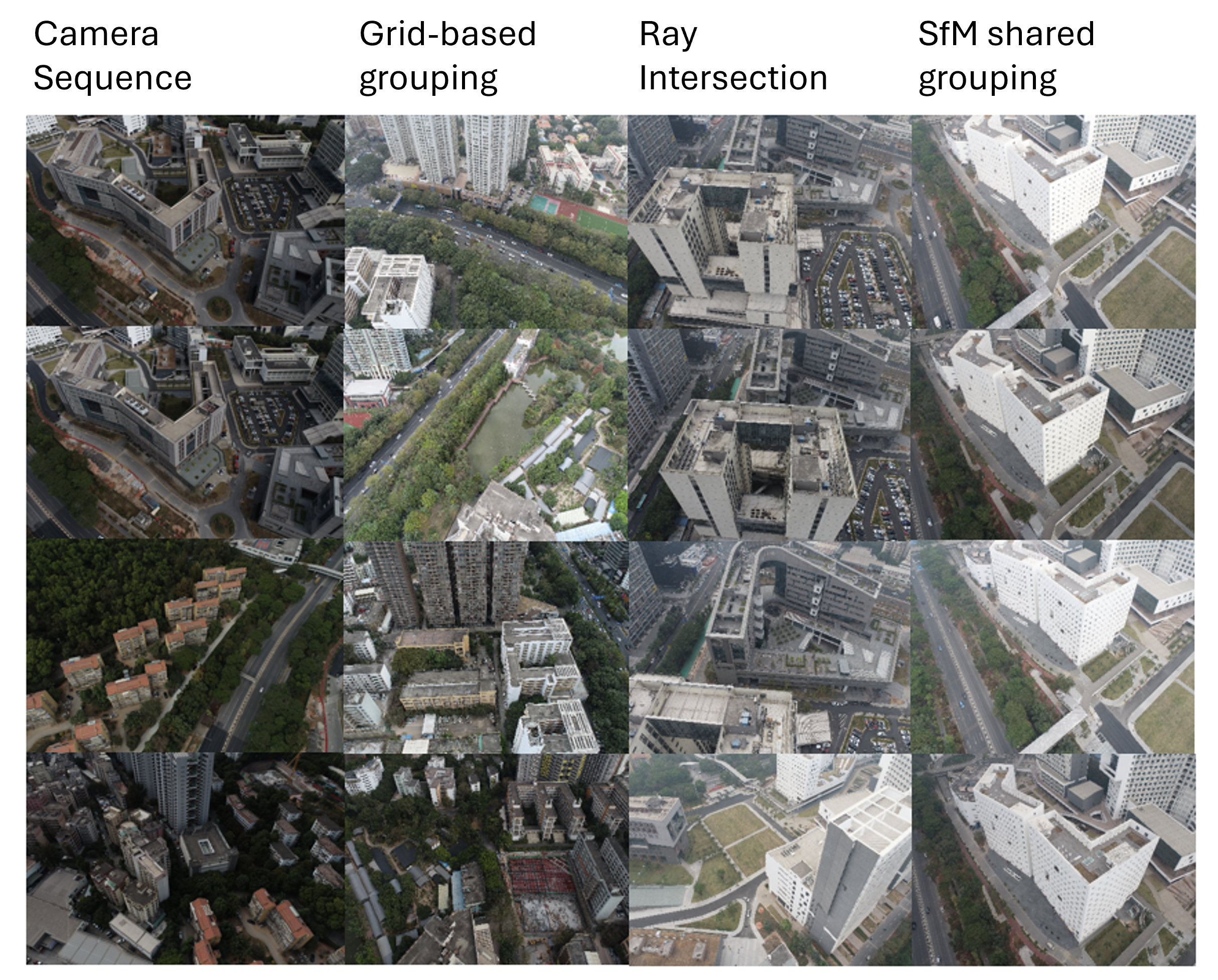}
    \caption{Qualitative comparison of clustering methods for aerial image grouping. Each column represents a method: (a) Camera Sequence groups images with overlap in scenes 1 and 2, but misses 3 and 4. (b) Grid-Based grouping overlaps scenes 1 and 3, missing others. (c) Ray Intersection captures overlap in scenes 1, 2, and partly 3, but not 4. (d) SfM Shared grouping achieves high overlap across all scenes, demonstrating superior performance.}

    \label{fig:grouping_approaches_qualitative}
\end{figure*}

\begin{figure*}
    \centering
    \includegraphics[scale = 0.6]{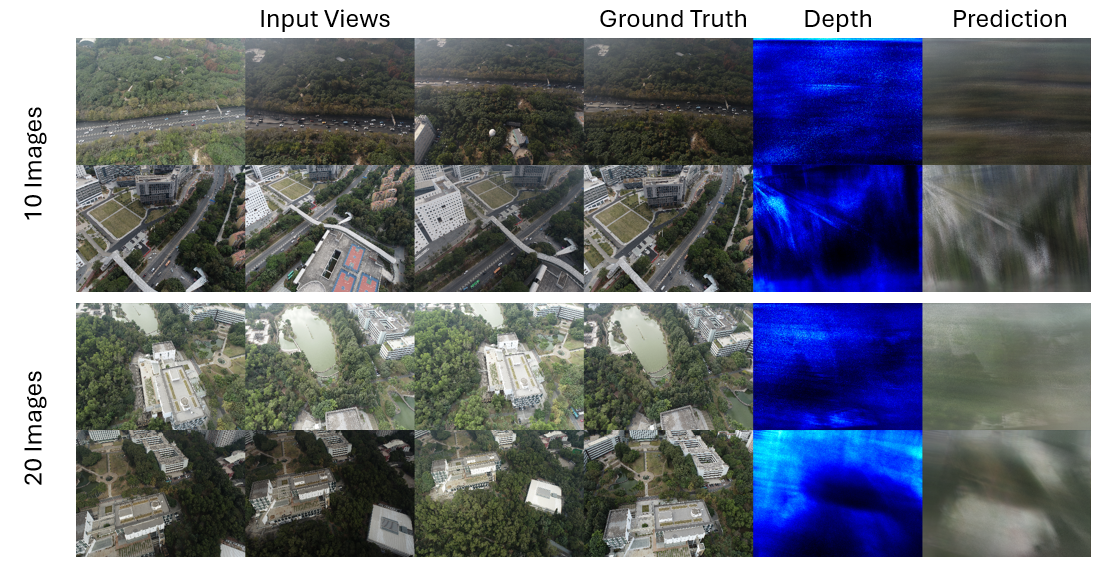}
    \caption{Qualitative comparison of models trained with 10 images per cluster versus 20 images per cluster using the SfM shared points method on a campus scene. Reducing the cluster size from 20 to 10 demonstrates marginally improved visual quality. The first row represents fine-grained predictions, while the second row shows coarse-grained predictions. Columns depict Input Views, Ground Truth, Depth, and Predictions.}

    \label{fig:20_vs_10_clustering}
\end{figure*}

\section{Experimental Setup}

\textbf{Dataset: } For our experimental analysis, we focus exclusively on the Campus scene from the UrbanScene3D \cite{lin_capturing_2022} dataset. This scene spans an area of $1.3 \times 10^{6} m^{2}$ and includes 178 objects, providing diverse urban structures for evaluation.

\textbf{Metric: } In evaluating our model, we will apply a combination of quantitative metrics and qualitative assessments. Quantitatively, we will utilize the Peak Signal-to-Noise Ratio (PSNR) to measure the fidelity of the reconstructions against the corrupting noise. Our approach will include visual inspections to assess the realistic rendering of the scenes. 

\textbf{Comparison: } Our analysis involves comparing the performance of PixelNeRF on the real dataset with its performance on an augmented dataset that combines real and synthetic data. To achieve this, we first evaluate PixelNeRF's performance on the real dataset alone, ensuring that the data is curated effectively. We identify the best-performing approach using the proposed four preprocessing methods described in Section \ref{sec:preprocess}. Once this baseline is established, we integrate augmentations generated through Aug3D, employing the two augmentation strategies detailed in Section \ref{sec:augmentation}, and compare the results to assess the impact of augmentation.

\textbf{Compute Setup: } PixelNeRF \cite{yu2020pixelnerf} is run with 256 hidden layers and fixed encoder weights, adhering to its default configuration to meet low computational requirements. We utilize two 32GB Tesla V100 GPUs to evaluate the real Campus dataset. Grid-based augmentation, combined with the real dataset, is processed on a single 24GB NVIDIA RTX 3090 Ti. All other experiments are conducted using a 10GB NVIDIA RTX 3080, ensuring consistency across setups where applicable.

\section{Results}
\label{sec:results}

\textbf{Evaluating Data Curation Methods:} Experiments on the Campus scene from UrbanScene3D \cite{rematas_urban_2022} demonstrate that \textit{SfM shared grouping} out of the methods mentioned in Section \ref{sec:preprocess} achieves the best performance for Generalizable Novel View Synthesis using PixelNeRF \cite{yu2020pixelnerf}. Using input images set to 3, a cluster size of 20, and Peak Signal-to-Noise Ratio (PSNR, higher the better) as the evaluation metric, \textit{SfM shared grouping} attains the highest PSNR of 20.03 and an average PSNR of 14.6, outperforming \textit{Camera sequence grouping} and \textit{Grid-based grouping}, with PSNR values of 9.7 and 12.2, respectively as shown in Table \ref{tab:clustering_methods}.
\begin{table}[h]
\centering
\caption{Performance of Different Clustering Methods}
\begin{tabular}{lccc}
\toprule
\textbf{Method} & \textbf{Best PSNR$\uparrow$} & \textbf{Low PSNR$\uparrow$} & \textbf{Avg. PSNR$\uparrow$} \\ 
\midrule
Sequence grouping & 9.7  & 0.0  & 3.5 \\ 
Grid-Based grouping & 12.2 & 0.0  & 4.6 \\ 
Ray intersection & 13.6 & 0.0  & 9.9 \\ 
SfM shared grouping & \textbf{20.03} & \textbf{10.9} & \textbf{14.6} \\ 
\bottomrule
\end{tabular}
\label{tab:clustering_methods}
\end{table}
Qualitative results in Fig. \ref{fig:grouping_approaches_qualitative} confirm that \textit{SfM shared grouping} provides better visual correspondence and hence leads to stable training performance. 
Reducing the cluster size from 20 to 10 further improves PSNR to 22.94, additionally highlighting the importance of high overlap within input clusters for reconstruction fidelity, also shown with qualitative results in Fig. \ref{fig:20_vs_10_clustering}.

\textbf{Baseline Performance: } Table \ref{table:results} details the results for the real and augmented datasets. For the real dataset, using a cluster size of 20 images, we observe a slight decline in PSNR as the number of input views increases. Specifically, the PSNR decreases from 20.03 for 3 input views to 19.59 for 9 input views. This trend suggests that while additional views provide more information, they may also introduce noise or redundancy that hinders GNVS performance.

\textbf{Aug3D + Real vs Real dataset: } The synthetic dataset, reconstructed using \textit{Grid Sampling} and \textit{Semantic Plane Fitting}, achieves PSNR values of 29.12 and 28.79, respectively, with 3 input images, a cluster size of 20. Augmenting the real dataset with these under the same parameters yields the best PSNR of 21.80 for the \textit{Semantic} approach, slightly surpassing \textit{Grid Sampling} at 21.67. These results validate the effectiveness of the Aug3D dataset in enhancing GNVS performance.

\section{Discussion}

This work demonstrates the potential of feed-forward Generalizable Novel View Synthesis (GNVS) models like PixelNeRF for large-scale outdoor scenes, exemplified by the UrbanScene3D dataset. To address the need for a dataset curation pipeline, we proposed four clustering strategies, identifying \textit{SfM shared grouping} as the most effective. Reducing the cluster size further improved performance, highlighting the critical role of high-view overlap. Additionally, our Aug3D augmentation method, which generates synthetic views through \textit{Grid Sampling} and \textit{Semantic Plane Fitting}, boosted GNVS performance when integrated with real data. Despite these advances, challenges remain, including mitigating noise from additional input views and ensuring scalability to diverse datasets and models, pointing to future directions in adaptive clustering and semantic-driven 3D augmentation.

\begin{table}[t]
\centering
\caption{Results for various datasets}
\begin{tabular}{l l l}
\toprule
\textbf{Dataset} & \textbf{Configuration} & \textbf{Best PSNR $\uparrow$} \\ 
\midrule
\textbf{Real Dataset (Baseline)} & Input views 3 & 20.03 \\ 
& Input views 6 & 19.95 \\ 
& Input views 9 & 19.59 \\ 
\midrule
\textbf{Synthetic Dataset (ours)} & Grid Sampling & 29.12 \\ 
& Semantic Plane Fitting & 28.79 \\ 
\midrule
\textbf{Aug3D (ours + baseline)} & Grid & 21.67 \\ 
& Semantic & 21.80 \\ 
\bottomrule
\end{tabular}
\label{table:results}
\end{table}

\bibliographystyle{splncs04}
\bibliography{references_aug3D}

\begin{figure*}
    \centering
    \includegraphics[scale = 0.6]{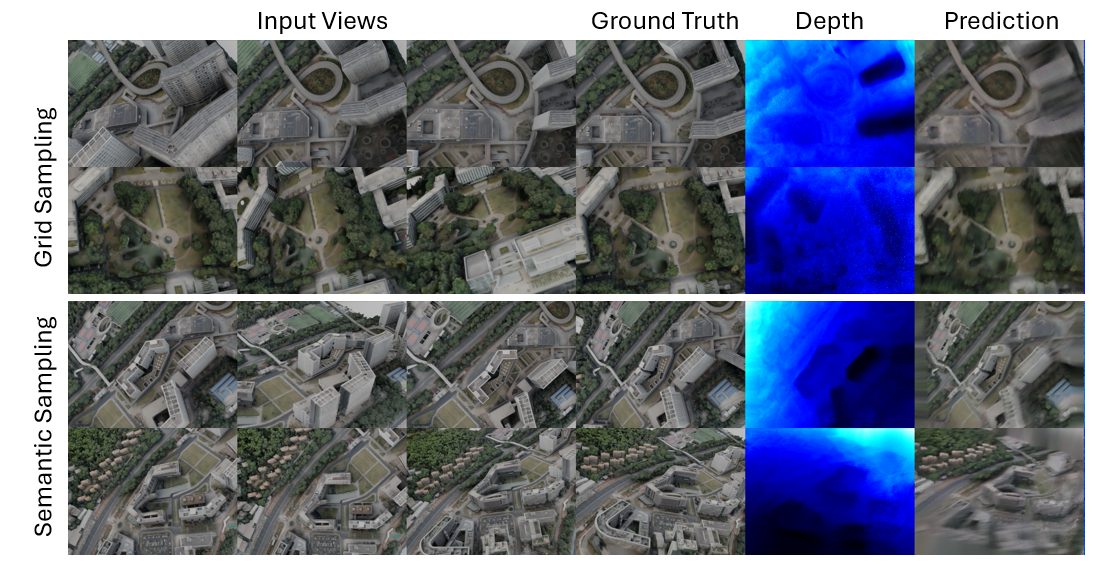}
    \caption{Qualitative comparison of PixelNeRF trained exclusively on synthetic datasets generated using grid sampling versus semantic sampling methods on the UrbanScene3D \textit{Campus scene}. The first row represents fine-grained predictions, while the second row shows coarse-grained predictions. }
    \label{fig:grid_vs_semantic_qualitative_campus}
\end{figure*}

\begin{figure*}
    \centering
    \includegraphics[scale = 0.6]{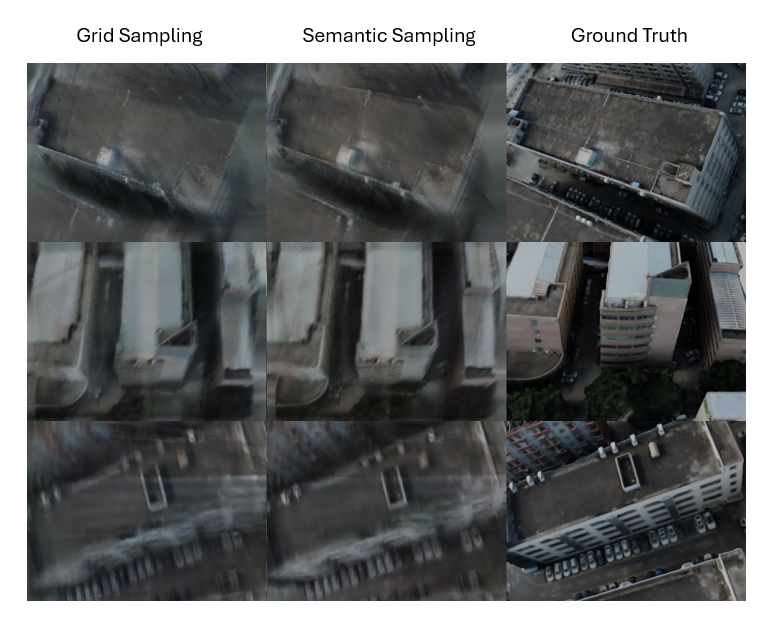}
    \caption{Qualitative comparison of PixelNeRF trained exclusively on synthetic datasets generated using grid sampling versus semantic sampling methods on the UrbanScene3D \textit{residence scene}, showcasing coarse predictions.}
    \label{fig:grid_vs_semantic_qualitative_residence}
\end{figure*}

\appendix

\section*{A. Other Semantic Sampling Method}

In addition to the geometric plane fitting approach, we experimented with a second semantic sampling method using the Segment Anything Model (SAM) to detect buildings from a top-down view. While SAM showed promise, it was found to be highly sensitive to shadows, resulting in inconsistencies in detecting building structures. Comparatively, the geometric plane fitting method yielded more reliable and accurate results, further emphasizing its suitability for generating semantically meaningful views in diverse lighting conditions.

\section*{B. Additional Qualitative Results}


To further illustrate the effectiveness of the proposed Aug3D augmentation strategies, we provide qualitative comparisons of the reconstructed scenes using \textit{Grid Sampling} and \textit{Semantic Plane Fitting}.

Figure \ref{fig:grid_vs_semantic_qualitative_campus} showcases the qualitative results on the UrbanScene3D Campus scene with 3 input images, comparing models trained exclusively on synthetic datasets generated via grid sampling versus semantic sampling methods. Notably, the semantic sampling approach demonstrates improved reconstruction fidelity, with sharper edges and more accurate structural details, particularly in regions with complex geometries. 

In Figure \ref{fig:grid_vs_semantic_qualitative_residence}, we extend this analysis to the Residence scene, evaluating the same augmentation techniques. Similar trends are observed, with semantic sampling outperforming grid sampling in preserving finer scene details and mitigating artifacts. The results underline the potential of semantic-driven augmentation to enhance the diversity and quality of synthetic datasets, thereby benefiting GNVS training.

These qualitative evaluations, along with our experiments, reinforce the quantitative findings presented in Section \ref{sec:results}, validating the advantages of integrating semantic-driven synthetic views into the GNVS pipeline. Future work can explore enhancing SAM's robustness to lighting variations or combining its capabilities with geometric methods for more versatile augmentation strategies.

\end{document}